\documentclass[letterpaper, 10 pt, conference]{ieeeconf}  

\IEEEoverridecommandlockouts                              

\usepackage{graphics} 
\usepackage{epsfig} 
\usepackage{mathptmx} 
\usepackage{times} 
\usepackage{amsmath} 
\usepackage{amssymb}  
\usepackage{url}
\usepackage{xcolor}
\usepackage[normalem]{ulem}
\usepackage{booktabs}

\title{\LARGE \bf
Vogtareuth Rehab Depth Datasets: Benchmark for Marker-less Posture Estimation in Rehabilitation
}

\author{Soubarna Banik$^{1}$, Alejandro Mendoza Garc\'{i}a$^{2}$, Lorenz Kiwull$^{3}$, Steffen Berweck$^{4}$, and Alois Knoll$^{1}$
\thanks{{\small \textit{Acknowledgment: The authors acknowledge the financial support by the Federal Ministry of Education and Research of Germany and the children and their parents for participating in the dataset.}}}
\thanks{$^{1}$Soubarna Banik and Alois Knoll are with Department of Informatics, Technical University of Munich, Germany
        {\tt\small soubarna.banik@tum.de,knoll@in.tum.de}}%
\thanks{$^{2}$Alejandro Mendoza Garc\'{i}a is with reFit Systems, Munich, Germany
        {\tt\small alejandro@refit-systems.com}}%
\thanks{$^{3}$Lorenz Kiwull is with Ludwig Maximilian University of Munich, Germany
        {\tt\small lorenz.kiwull@lmu.de}}%
\thanks{$^{4}$Steffen Berweck is with Clinic for Neuropediatrics and Neurorehabilitation, Epilepsy Center for Children and Adolescents, Sch\"{o}n Klinik Vogtareuth, Germany
        {\tt\small sberweck@schoen-klinik.de}}%
}

\usepackage{tikz}

\newcommand\copyrighttext{%
  \footnotesize \textcopyright 2021 IEEE. Personal use of this material is permitted.  Permission from IEEE must be obtained for all other uses, in any current or future media, including reprinting/republishing this material for advertising or promotional purposes, creating new collective works, for resale or redistribution to servers or lists, or reuse of any copyrighted component of this work in other works.}
\renewcommand\copyrightnotice{%
\begin{tikzpicture}[remember picture,overlay]
\node[anchor=south,yshift=10pt] at (current page.south) {\fbox{\parbox{\dimexpr\textwidth-\fboxsep-\fboxrule\relax}{\copyrighttext}}};
\end{tikzpicture}%
}

\newcommand\blfootnote[1]{
	\begingroup
	\renewcommand\thefootnote{}\footnote{\vspace{-0.4cm} #1}
	\addtocounter{footnote}{-1}
	\endgroup
}
\begin{document}

\maketitle
\blfootnote{\copyrightnotice}
\thispagestyle{empty}
\pagestyle{empty}

\begin{small}
\noindent To appear in {\it Proc.\ IEEE EMBC 2021, November 01-05, 2021.}\\
\end{small}

\begin{abstract}
Posture estimation using a single depth camera has become a useful tool for analyzing movements in rehabilitation.
Recent advances in posture estimation in computer vision research have been possible due to the availability of large-scale pose datasets.
However, the complex postures involved in rehabilitation exercises are not represented in the existing benchmark depth datasets.
To address this limitation, we propose two rehabilitation-specific pose datasets containing depth images and 2D pose information of patients, both adult and children, performing rehab exercises.
We use a state-of-the-art marker-less posture estimation model which is trained on a non-rehab benchmark dataset.
We evaluate it on our rehab datasets, and observe that the performance degrades significantly from non-rehab to rehab, highlighting the need for these datasets.
We show that our dataset can be used to train pose models to detect rehab-specific complex postures.
The datasets will be released for the benefit of the research community.
\end{abstract}

\section{INTRODUCTION}

Posture estimation is an important tool for gait analysis.
Among the vision-based approaches for posture estimation, marker-less methods are gaining momentum, as they are user-friendly and non-invasive in nature.
Though the marker-based methods~\cite{ref_url_vicon,ref_url_optitrack} are more accurate, the space requirement, installation overhead, complex operation process, and cost make their widespread use in rehab clinics difficult.

The introduction of Microsoft's Kinect and its marker-less skeleton tracking algorithm \cite{kinect} paved the way for an inexpensive alternative for the clinics.
Multiple studies have been conducted employing Kinect for movement analysis~\cite{kinectrehab1,kinectrehab2}.
However, studies such as \cite{kinectrehab3,kinectrehab4} find that Kinect V2 system is reliable for assessing simple movements, but fails for complex postures such as double leg squat, hip abduction, lunge, or for movements with small amplitudes.

With the recent advances in deep-learning based marker-less posture estimation methods complex postures such as cycling, bending, etc. can now be detected.
This has been particularly possible because of the availability of large-scale human posture datasets such as COCO \cite{coco} and Human3.6M \cite{human36m}.
As deep-learning based approaches are data-driven, their performance depends on the training data.
However, the benchmark datasets are generic and mostly contain images of adults.
Rehabilitation scenarios involving patients of varied ages are underrepresented there.
Also, the color images raise privacy concerns in the medical domain.
Researchers in rehabilitation have developed their own pose datasets in past \cite{HPERehabPickCup}, \cite{hesse2018computer}, which are limited to specific movements and poses such as `Picking up a cup' motion \cite{HPERehabPickCup} and lying poses of infants and adults \cite{hesse2018computer}.
Due to the lack of large-scale rehab-specific pose datasets, only a limited number of work has adapted the state-of-the-art pose models in rehab \cite{HPERehabPickCup}, \cite{HPERehabLying}.

In this paper, we propose two newly created rehab depth datasets labeled with 2D poses.
Our datasets have varied samples of adults and children, different genders, and also instances of limited range of motion for some joints.
We evaluate a state-of-the-art posture estimation algorithm on our datasets.
We show that the performance of model trained on generic pose dataset does not translate to rehabilitation-specific scenarios, justifying the need for these datasets.

\section{Datasets}
As privacy is an obligation in children's rehabilitation, we collected only depth images.
We created two depth datasets, namely Vogtareuth Rehab (VtR) and Vogtareuth Rehab -- Optitrack (VtR-O), containing images of children and adults performing rehabilitation exercises.
The two datasets differ in the complexity of the poses and the type of motion-capture systems used for annotation.

\subsection{Data collection setup}
A data collection setup was built in Sch\"{o}n Klinik, Vogtareuth with six infrared Optitrack cameras~\cite{ref_url_optitrack} and a Kinect V2 camera.
A program called MoveLab from reFit Systems~\cite{ref_url_refit} was used to record and synchronize the data from both Kinect and Optitrack. 
The depth data from Kinect V2 was clamped at 4 meters and was compressed using a lossless compression algorithm \cite{wilson}.

\subsection{Vogtareuth Rehab (VtR) Dataset}
For this dataset, we collected data of 9 healthy users including 7 children and 2 adults, of male and female genders, while they played reFit Gamo games~\cite{gamo} for rehabilitation.
The actions involved in the games are sideways movement of the trunk in the sagittal plane, arm abduction, arm flexion, knee raise, bending, and some free movements.
Every $15^{th}$ frame was processed and 2D poses derived by Kinect's skeleton detection algorithm were extracted.
Altogether 24,742 depth images and 15 joint locations were extracted from the raw data, out of which 3712 images were reserved for the test set.
We discarded the frames where the number of noisy joints, i.e. joints estimated outside the body, exceeded two.

\begin{figure}[t]
\centering
\includegraphics[width=\columnwidth]{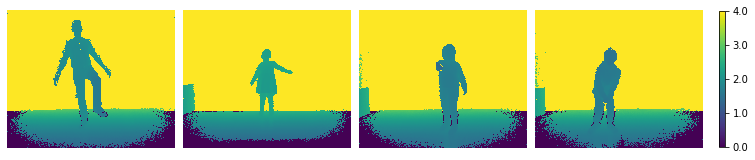}
\caption{Sample depth images from VtR dataset, showing following actions: raising knee, arm abduction, arm flexion and squat.} \label{fig:sk1}
\end{figure}
\begin{figure}[t]
\centering
\includegraphics[width=\linewidth]{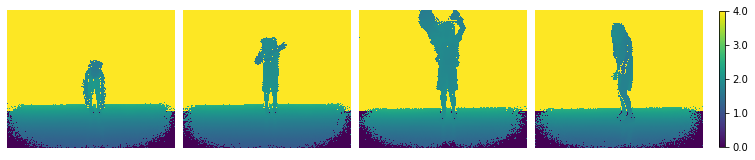}

\caption{Sample depth images from VtR-O dataset, showing following actions: touching toe, crisscrossing arms, raising both arms upwards, and turning 360$^{\circ}$.} \label{fig:sk2}
\end{figure}

\subsection{Vogtareuth Rehab -- Optitrack (VtR-O) Dataset}
\label{sec:VRO}
In addition to VtR, a smaller dataset of complex postures was created.
A series of actions such as raising both arms upwards, crisscrossing arms while stretched out in front, turning 360$^{\circ}$, and bending down to touch the toes were specifically selected, where we observed Kinect to fail in detecting the correct pose. 
We used Optitrack \cite{ref_url_optitrack}, a marker-based motion capture system for recording precise joint locations for the postures. 
We record 2 children with restricted motions in one arm and one leg respectively, as they were instructed by a physiotherapist to perform the actions.

We placed 27 markers on the upper body following the conventional upper body markerset of Optitrack \cite{ref_url_optitrack}.
Six additional markers were placed on the lower body - on the knees, ankles, and foot tips.
This hybrid markerset was chosen as Optitrack's full body markerset has 39 markers which are difficult for the children to wear.
Optitrack recorded the precise 3D locations of 13 upper body joints and the 6 additional markers for the lower body.
Out of the 19 Optitrack joints, 15 joints as in VtR were extracted.
The leg markers were labeled manually with the joint names.
Fig. \ref{fig:sk2} shows sample images from this dataset.
As can be seen in the figure, Optitrack's infrared markers create holes in the depth image.
Also, noisy protrusions are detected around the markers.
This is caused by the interference between Kinect V2 and Optitrack.

We synchronized the data from the two sources using the timestamp.
Unlike VtR, we processed every frame for which skeletons from both devices were available.
The marker holes were removed from the depth images using a median filter.
We did not discard frames with Kinect's noisy skeleton as our objective is also to compare Kinect's performance with that of Optitrack.
Optitrack's 3D skeleton was first transformed to Kinect's coordinate frame by subtracting Kinect's position and then transformed to Kinect's 2D depth space using Kinect's coordinate mapping algorithm.
Altogether 2145 depth images and 2D joint locations from both Kinect and Optitrack were recorded.

\section{Experiments}
In this section, we evaluate an off-the-shelf model, trained using a generic pose dataset, on the rehabilitation dataset VtR.
The objective is to analyze the performance shift, if any, from non-rehab to rehab.
Additionally, we also highlight the shortcomings of Kinect-annotated data.
We do so by training a model with the VtR dataset and evaluating it against Optitrack's ground truth in the VtR-O dataset.


\subsection{Pose Estimation Model}
There exist multiple deep learning-based 2D pose estimation methods that perform well on benchmark pose datasets.
We choose a ResNet-based method \cite{poseresnet}, hereafter referred to as Pose-ResNet.
It has a relatively simple architecture in contrast to the other complex multi-stage designs \cite{cpm,stackedhg} and yet has comparable performance.
For our experiment, we use the Pose-ResNet-50.
Pose-ResNet originally takes a 3 channel RGB image as input.
As depth images contain only one channel, we convert it to a 3 channel input by copying it to the rest of the channels. 
Moreover, we use 128 filters in the deconvolutional layers instead of 256.
The network produces a 2D heatmap for every joint as output.
The heatmap indicates the location of the corresponding joint in the input scene.
The joint location is extracted from the heatmap by selecting the pixel with maximum value.

\subsection{Training Details}
State-of-the-art pose models, including Pose-ResNet, are mostly trained on color images.
Hence, we need to retrain the model on depth images.
For the generic dataset, we choose a benchmark pose depth dataset, named ITOP \cite{itop}.
It contains poses of 20 adults performing 15 action sequences, which are not specific to rehabilitation.
The images are annotated with 15 body joints extracted using \cite{kinect}, similar to VtR.
We use only clean and human-approved images from the \textit{side view} set of ITOP \cite{itop}.
Additionally, we also train Pose-ResNet on our VtR dataset to compare its performance with Optitrack's data.
The models trained on ITOP and VtR are named as Pose-ResNet-ITOP and Pose-ResNet-VtR respectively.
The ITOP dataset is split into training, validation and test sets of 13623, 3406 and 4606 images, and VtR into 17244, 3786 and 3712 images respectively.

The training procedure for both models is the same.
Data augmentation techniques such as random cropping 
and flipping are applied.
The depth values in the image are normalized to [0,1], using min-max normalization, where the minimum and maximum depths are set to 0 and 4 meters respectively.
Finally, the image is resized to $224\times224$.
Target heatmaps of $64\times64$ size are generated for each joint by applying a 2D Gaussian centered on the joint's ground truth location.
We use the Adam optimizer, a learning rate of 2.5$\times$10$^{-4}$, a dropout rate of 0.5, and mean squared error loss between the predicted and target heatmaps.
Both Pose-ResNet-ITOP and Pose-ResNet-VtR are finetuned for 10 epochs on the training set of the respective datasets.


\begin{table}[t]
\caption{PCK and PCKh of Pose-ResNet-ITOP on ITOP and VtR.}
\label{tab1}
\begin{center}
\begin{tabular}{l|c|c}
\hline
\multicolumn{1}{c|}{Dataset}	& PCK  & PCKh \\
\hline
 ITOP &            89.21 &      89.12 \\
 VtR/All &          41.08 &      22.31  \\
 VtR/Adult &        48.73 &      28.96  \\
 VtR/Child &        38.27 &      19.87  \\
\hline
\end{tabular}
\end{center}
\end{table}

\begin{figure}[t]
\centering
\includegraphics[width=0.9\columnwidth]{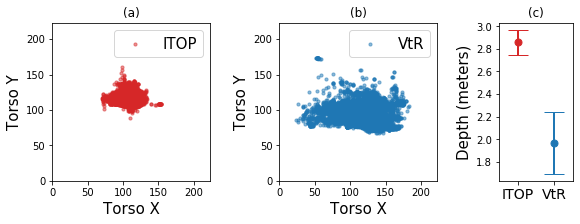}
\caption{Visualization of torso variability. Torso pixel position in (a) ITOP (b) VtR dataset. Each point represents an image. (c) Mean and standard deviation of torso depth for both datasets. } \label{fig:torso}
\end{figure}

\begin{figure}[t]
\centering
\includegraphics[width=0.9\linewidth]{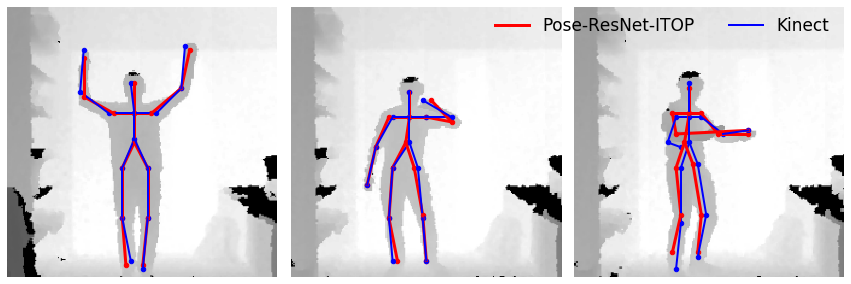}
\caption{Sample results of Pose-ResNet-ITOP from ITOP dataset.} \label{fig:results}
\end{figure}

\begin{figure}[t]
\centering
\includegraphics[width=0.9\linewidth]{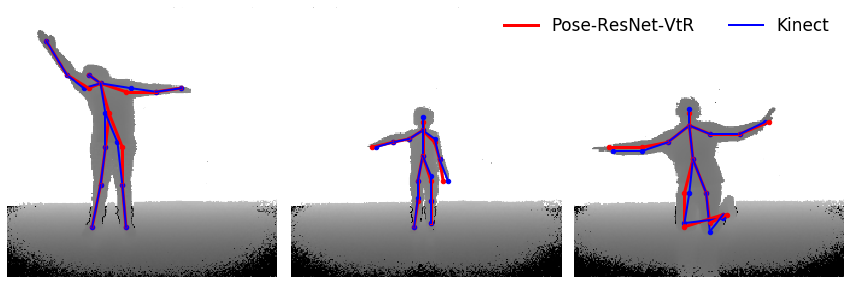}
\caption{Sample results of Pose-ResNet-VtR from VtR dataset, showing both adult and children.} \label{fig:results1}
\end{figure}

\subsection{Evaluation Metrics}
The model is evaluated using the Percentage of Correct Keypoints (PCK) metric \cite{cpm, stackedhg} with a variable threshold.
PCK is the percentage of correctly detected joints.
A detected joint is considered correct if the normalized distance between the predicted and the true joint is within a threshold value.
The distance is normalized by the torso size.
Additionally, we use PCKh metric \cite{stackedhg}, where the distance is normalized by head size.
Both metrics use adaptive thresholding based on the subject's torso or head size and hence can deal with the variation in size and scale of people.

\subsection{Evaluation on ITOP and VtR datasets}

Table \ref{tab1} reports the PCK and PCKh scores of Pose-ResNet-ITOP at $0.5$ threshold on the test sets of ITOP and VtR.
The model performs well on the ITOP test set in both metrics.
However, there is a significant drop in performance, when evaluated on the VtR dataset, from 89.21\% to 41.08\% in PCK and worse for PCKh.
To investigate further, we evaluate the performance on adults and children separately, and observe that the model performs worse for children.
The training dataset of ITOP does not contain any samples of children, which explains this behavior.

Additionally, we observe that the subjects in ITOP are standing mostly in the image center, and the movements are also restricted to a small region in the depth or Z axis.
On the contrary, the VtR dataset has higher variability in terms of the subjects' position in X, Y and Z axis.
This can be seen from the torso positions and the corresponding depth values reported in Fig.~\ref{fig:torso} for both datasets.
Due to the higher variance in size, scale and position of the subjects, VtR is more challenging, and hence the performance is affected.

The PCKh score of Pose-ResNet-ITOP on VtR dataset is much lower than the corresponding PCK (refer Table \ref{tab1}).
As head size is smaller than torso size, the effective threshold in PCKh is smaller compared to PCK, resulting in lower PCKh scores in general.
However, the drop in score is negligible for ITOP.
We observe that in the ITOP dataset the position of the neck is midway between the head and torso joints, as can be seen in Fig. \ref{fig:results}.
This leads to the mean head and torso sizes becoming very similar, with a difference of only 0.13 pixels.
On the other hand, in VtR, this difference is 12.97 pixels.
As seen in Fig. \ref{fig:results1}, our annotation in the VtR dataset agrees more with the anatomical structure of the human body and so is more relevant for possible rehabilitation analysis.

When Pose-ResNet is trained on the generic pose dataset, it cannot perform well on VtR, which contains complex and varied postures specific to rehabilitation.
Pose-ResNet-VtR which is trained on the VtR dataset achieves 93.90\% and 86.03\% in PCK and PCKh metric respectively on the VtR test set.
This implies that deep-learning models trained using our rehabilitation-specific dataset can be effectively utilized in rehabilitation scenarios.

\begin{figure}[t]
\centering
\includegraphics[width=0.9\linewidth]{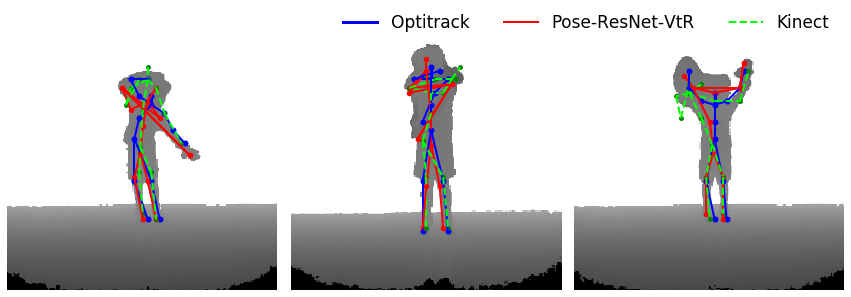}
\caption{Sample results of Pose-ResNet-VtR and Kinect's algorithm from VtR-O dataset} \label{fig:results2}
\end{figure}

\begin{figure}[t]
\centering
\includegraphics[width=\columnwidth]{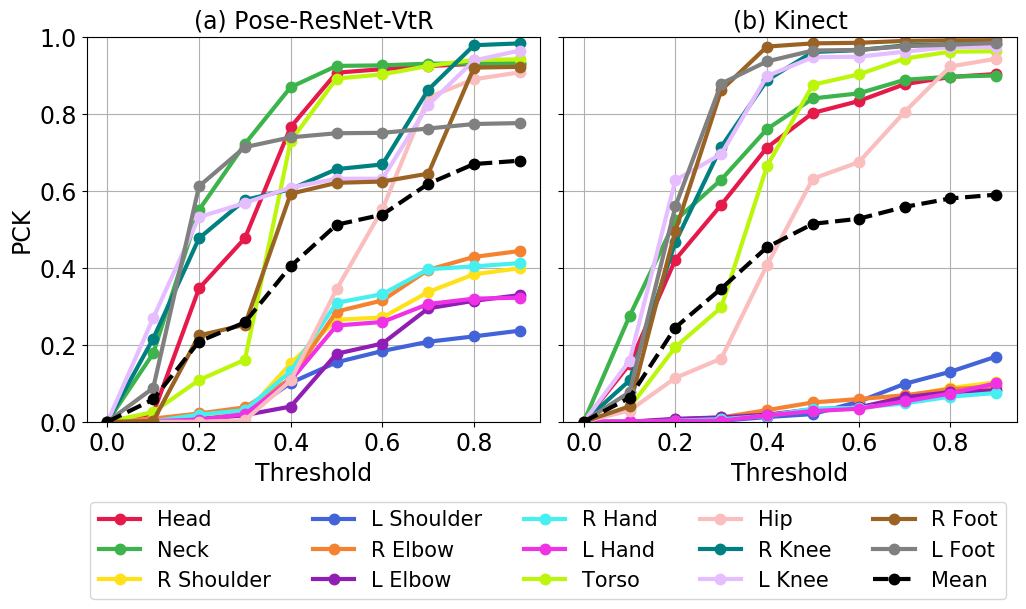}
\caption{PCK score on VtR-O: Mean and joint level PCK of Model and Kinect against Optitrack ground truth.} \label{fig:plot2}
\end{figure}

\subsection{Evaluation on VtR-O dataset}

We evaluate the Pose-ResNet-VtR model also on the VtR-O dataset.
As explained in section \ref{sec:VRO}, VtR-O contains complex postures, where we have seen Kinect's algorithm to fail specifically.
In this study, we aim to analyze the performance of models trained on Kinect's skeletal data with respect to Optitrack's precise annotation.
For this, we consider Pose-ResNet-VtR and also Kinect's algorithm itself.

Fig. \ref{fig:results2} shows some sample results from VtR-O dataset.
The mean and joint-level PCK scores of both models are shown in Fig. \ref{fig:plot2} for a range of threshold values from $0.0$ to $1.0$.
With increasing threshold, the margin of error is higher and so is the PCK score.
Complex poses in the VtR-O dataset such as raising arms and standing sideways, together with the subjects' size make the pose detection task challenging.
This is reflected by the large degradation of the mean score of Pose-ResNet-VtR on VtR-O: the performance drops from 93.90\% in VtR test-set to 49.60\% for 0.5 threshold.
The mean score of Kinect's model is 51.64\% for these complex poses.

At joint-level, Kinect's algorithm performs poorly for the upper-limb joints (see PCK curves for hands, elbows and shoulders in Fig. \ref{fig:plot2}), as they are too close to the body in VtR-O dataset.
Also, both models confuse the head with the hand, when the hand is too close to the head.
This is shown in the right-most image in Fig. \ref{fig:results2}.
 
As Optitrack's ground truth skeleton is more precise, we plan to extend the VtR-O dataset in the future, so that pose estimation models for rehabilitation can be trained using this dataset.
With more accurate ground truth for complex poses, the performance of the models will increase further.
 
\section{CONCLUSIONS}

We present two depth datasets, namely Vogtareuth Rehab (VtR) and Vogtareuth Rehab -- Optitrack (VtR-O) for 2D poses.
The datasets contain complex poses specific to rehabilitation exercises and are diverse in terms of age, size, gender, position and also range of motions of the subjects. 
We evaluate a state-of-the-art deep-learning based pose estimation method which is trained on a generic pose depth dataset on our datasets.
We show that the performance degrades because of the domain shift from non-rehab to rehab, showing the need for a rehab-specific dataset.
In the future, we want to extend the VtR-O dataset with more complex poses, for example involving wheelchairs, so that Optitrack's accurate skeletal data can be used to train and improve deep-learning based marker-less methods.





%
%
%



\end{document}